\documentclass[twoside,11pt]{article}
\usepackage{jmlr2e}
\ShortHeadings{Torchbearer}{E. Harris, M. Painter and J. Hare} 

\usepackage[english]{babel}
\usepackage[utf8]{inputenc}

\usepackage{hyperref}
\hypersetup{
    colorlinks=true,
    citecolor=blue,
    linkcolor=blue,
    filecolor=blue,      
    urlcolor=blue,
}
\usepackage{listings}
\usepackage[symbol]{footmisc}

\DeclareFixedFont{\ttb}{T1}{txtt}{bx}{n}{8} 
\DeclareFixedFont{\ttm}{T1}{txtt}{m}{n}{8}  

\usepackage{color}
\definecolor{deepblue}{rgb}{0,0,0.5}
\definecolor{deepred}{rgb}{0.6,0,0}
\definecolor{deepgreen}{rgb}{0,0.5,0}

\usepackage{parcolumns}

\newcommand\torchbearer{\texttt{torch\-bearer}}
\newcommand\version{0.2.1}

\newcommand\pythonstyle{\lstset{
language=Python,
commentstyle=\ttm,
basicstyle=\ttm,
otherkeywords={self},             
keywordstyle=\ttb\color{deepblue},
emph={MyClass,__init__},          
emphstyle=\ttb\color{deepred},    
stringstyle=\color{deepgreen},
frame=single,                         
showstringspaces=false
}}

\lstnewenvironment{python}[1][]
{
\pythonstyle
\lstset{#1}
}
{}

\newcommand\pythoninline[1]{{\pythonstyle\lstinline!#1!}}

\title{Torchbearer: A Model Fitting Library for PyTorch}

\author{\name Ethan Harris\footnotemark[1] \email ewah1g13@ecs.soton.ac.uk\\ 
\name Matthew Painter\footnotemark[1] \email mp2u16@ecs.soton.ac.uk\\ 
\name Jonathon Hare \email jsh2@ecs.soton.ac.uk\\ 
\addr Vision, Learning and Control Group\\ 
Department of Electronics and Computer Science\\ 
University of Southampton, United Kingdom}

\begin{document}

\maketitle

\begin{abstract}
We introduce \torchbearer{}, a model fitting library for \texttt{pytorch} aimed at researchers working on deep learning or differentiable programming. The \torchbearer{} library provides a high level metric and callback API that can be used for a wide range of applications. We also include a series of built in callbacks that can be used for: model persistence, learning rate decay, logging, data visualization and more. The extensive documentation includes an example library for deep learning and dynamic programming problems and can be found at \url{http://torchbearer.readthedocs.io}. The code is licensed under \href{https://opensource.org/licenses/MIT}{the MIT License} and available at \url{https://github.com/ecs-vlc/torchbearer}.
\end{abstract}


\footnotetext[1]{Contributed equally to this work and are listed in alphabetical order.}

\section{Introduction}

The meteoric rise of deep learning will leave behind a host of frameworks that support hardware accelerated tensor processing and automatic differentiation. We speculate that over time a more general characterization, differentiable programming, will take its place. This involves optimizing the parameters of some differentiable program through gradient descent in a process known as fitting. One library which has seen increasing use over recent years is \texttt{pytorch} \citep{paszke2017automatic}, which excels, in part, because of the ease with which one can construct models that perform non-standard tensor operations. This makes \texttt{pytorch} especially useful for research, where any aspect of a model definition may need to be altered or extended. However, as \texttt{pytorch} is specifically focused on tensor processing and automatic differentiation, it lacks the high level model fitting API of other frameworks such as \texttt{keras} \citep{chollet2015keras}, a library for training deep models with \texttt{tensorflow}. Furthermore, such libraries rarely support the more general case of differentiable programming.

We introduce \torchbearer{}, a \texttt{pytorch} library that aids research by streamlining the model fitting process whilst preserving transparency and generality. At the core of \torchbearer{} is a model fitting API where every aspect of the fitting process can be easily modified. We also provide a powerful metric API which enables the gathering of rolling statistics and aggregates. Finally, \torchbearer{} provides a host of callbacks that perform advanced functions, such as using \href{https://github.com/lanpa/tensorboardX}{\texttt{tensorboardx}} to log visualizations to \texttt{tensorboard} (included with \texttt{tensorflow} \citep{tensorflow2}) or \href{https://github.com/facebookresearch/visdom}{\texttt{visdom}}.

\section{Design}

The \torchbearer{} library is written in Python \citep{Python} using \texttt{pytorch}, \texttt{torch\-vision} and \texttt{tqdm} and depends on \texttt{numpy} \citep{oliphant2006guide}, \texttt{scikit-learn} \citep{pedregosa2011scikit} and \texttt{tensor\-boardx} for some features. The key abstractions in \torchbearer{} are trials, callbacks and metrics. There are three key principles which have motivated the design:
\begin{itemize}
	\item Flexibility: The library supports both deep learning and differentiable programming.
    \item Functionality: The library includes functions that make complex behavior available.
    \item Transparency: The API is simple and clear.
\end{itemize}

By virtue of these design principles, \torchbearer{} differs from other, similar libraries such as \href{https://github.com/pytorch/ignite}{\texttt{ignite}} or \href{https://github.com/pytorch/tnt}{\texttt{tnt}}. For example, neither library offers a wide range of built in callbacks for complex functions. Furthermore, both \texttt{ignite} and \texttt{tnt} use an event driven API for the model fitting which makes code less transparent and human readable.

\begin{figure}
\begin{minipage}{.45\textwidth}
\begin{python}[caption={Example callback}, captionpos=b, label=callback]
@callbacks.on_criterion
def l2_weight_decay(state):
  for W in state[MODEL].parameters():
    state[LOSS] += W.norm(2)
\end{python}
\end{minipage}
\hfill
\noindent\begin{minipage}{.45\textwidth}
\begin{python}[caption={Example metric definition}, captionpos=b, label=metric]
@metrics.default_for_key("acc")
@metrics.running_mean
@metrics.std
class CategoricalAccuracy...
\end{python}
\end{minipage}
\end{figure}

\subsection{Trial API}
The \texttt{Trial} class defines a \texttt{pytorch} model fitting interface built around \texttt{Trial.run(\ldots)}. There are also \texttt{predict(\ldots)} and \texttt{evaluate(\ldots)} methods which can be used for model inference or evaluating saved models. The \texttt{Trial} class also provides a \texttt{state\_dict} method which returns a dictionary containing: the model parameters, the optimizer state and the callback states which can be saved and later reloaded using \texttt{load\_state\_dict}.

\subsection{Callback API}
The callback API defines classes that can be used to execute functions at different points in the fitting process. The mutable state dictionary is an integral part of \torchbearer{} that is given to each callback and contains intermediate variables required by the \texttt{Trial}. This allows callbacks to alter the nature of the fitting process dynamically. Callbacks can be implemented as a decorated function through the decorator API, as shown in Listing \ref{callback}.

\subsection{Metric API}
The metric API uses a tree that enables data flow from one metric to a set of children. This allows for aggregates such as a running mean or standard deviation to be computed. Assembling these data structures can be difficult, and as such, \torchbearer{} includes a decorator API that simplifies construction. For example, in Listing \ref{metric} producing the standard deviation and running mean of an accuracy metric. The \texttt{default\_for\_key(\ldots)} decorator enables the metric to be referenced with a string in the \texttt{Trial} definition.

\section{Example: Generative Adversarial Networks (GANs)}
A GAN \citep{goodfellow2014generative} is a type of model that aims to learn a high quality estimation of an input data distribution. GANs comprise of two networks which are trained simultaneously but with opposing goals, the `generator' and the `discriminator'. To encourage the generator to produce samples that appear genuine, we use an adversarial loss. This is minimized when the discriminator predicts that generated samples are genuine. For the discriminator, we use a loss that maximizes the probability of correctly identifying real and fake samples. Implementing this requires forward passes from the generator and discriminator and separate backward passes for the loss of each network. Due to this complexity, training GANs is often challenging in typical frameworks since it requires a very flexible training loop, such as that of the \torchbearer{} \texttt{Trial}. In this section we will train a standard GAN with \torchbearer{} to demonstrate its effectiveness. 

\begin{figure}
\begin{python}[caption={GAN forward pass}, captionpos=b, label=forward]
class GAN(torch.nn.Module):
  def forward(real_imgs, state):
    z = (random sample of prior noise distribution)
    state[GEN_IMGS] = generator(z)
    state[DISC_GEN] = discriminator(state[GEN_IMGS])
    discriminator.zero_grad()

    state[DISC_GEN_DET] = discriminator(state[GEN_IMGS].detach())
    state[DISC_REAL] = discriminator(real_imgs)
\end{python}
\end{figure}

\begin{figure}
\begin{python}[caption=Loss computation, captionpos=b, label=loss]
@callbacks.add_to_loss
def gan_loss(state):
    fake_loss = adversarial_loss(state[DISC_GEN_DET], fake)
    real_loss = adversarial_loss(state[DISC_REAL], valid)
    state[G_LOSS] = adversarial_loss(state[DISC_GEN], valid)
    state[D_LOSS] = (real_loss + fake_loss) / 2
    return state[G_LOSS] + state[D_LOSS]
\end{python}
\end{figure}

To begin, the forward passes of both networks are combined in Listing \ref{forward}  with intermediate tensors stored in state. When \torchbearer{} is not passed a criterion, the base loss is automatically set to zero so that callbacks can add to it. We will use the \texttt{add\_to\_loss} callback shown in Listing \ref{loss} to combine the losses. The state keys are generated using \texttt{torchbearer.state\_key(key)} to prevent collisions. Note that the call to the underlying model does not automatically include the state dictionary, we will set the \texttt{pass\_state} flag in \texttt{Trial(\ldots)} to achieve this. Having created a data loader and optimizer using \texttt{pytorch}, we train the model with just two lines in listing \ref{training}.

\begin{figure}
\begin{python}[caption=Training, captionpos=b, label=training]
trial = Trial(GAN(), optimizer, metrics=["loss"], callbacks=[gan_loss], pass_state=True)
trial.with_train_generator(dataloader).run(epochs=200)
\end{python}
\end{figure}

\section{Project Management}
The \torchbearer{} library is licensed under \href{https://opensource.org/licenses/MIT}{the MIT License} and the most recent release (\version{}) is referenced on the \href{https://pypi.org/project/torchbearer/}{Python Package Index (PyPi)}. The code is hosted on \href{https://github.com/ecs-vlc/torchbearer}{GitHub}, see \href{https://github.com/ecs-vlc/torchbearer/blob/master/CHANGELOG.md}{CHANGELOG.md} for release information. The repository is actively monitored and we encourage users to raise issues requesting fixes or new functionality and to open pull requests for anything they have implemented.

\subsection{Continuous Integration}

To support usability, the library must be as stable as possible. For this reason we use continuous integration with \href{https://travis-ci.org/}{Travis CI} which tests all pull requests before they can be merged with the master branch. We also use \href{https://www.codacy.com/product}{Codacy} to perform automated code reviews which ensure that new code follows the PEP8 standard. In this way we ensure that the master copy of \torchbearer{} is always correctly styled and passes the tests.

\subsection{Example Library}

An effective way to improve usability is to provide examples of using the library for a range of problems. As such, we have added an \href{https://torchbearer.readthedocs.io/en/latest/examples/quickstart.html}{example library} which includes detailed examples showing how to use \torchbearer{} for various deep learning and differentiable programming models including: \href{https://torchbearer.readthedocs.io/en/latest/examples/gan.html}{GANs} \citep{goodfellow2014generative}, \href{https://torchbearer.readthedocs.io/en/latest/examples/vae.html}{Variational Auto-Encoders} \citep{kingma2013auto} and \href{https://torchbearer.readthedocs.io/en/latest/examples/svm_linear.html}{Support Vector Machines} \citep{cortes1995support}.

\section{Conclusion}

To summarize, \torchbearer{} is a library simplifies the process of fitting deep learning and differentiable programming models in \texttt{pytorch}. This is done without reducing transparency so that it is still useful for the purpose of research. Key features of \torchbearer{} include a comprehensive set of built in callbacks (such as logging, weight decay and model check-pointing) and a powerful metric API. The \torchbearer{} library has a strong and growing community on GitHub and we are committed to improving it wherever possible.

\bibliography{main.bib}

\end{document}